\documentclass{article}
\PassOptionsToPackage{numbers, compress}{natbib}
\usepackage[preprint]{neurips_2025}
\usepackage[utf8]{inputenc} 
\usepackage[T1]{fontenc}    
\usepackage{hyperref}       
\usepackage{url}            
\usepackage{booktabs}       
\usepackage{amsfonts}       
\usepackage{nicefrac}       
\usepackage{microtype}      
\usepackage{xcolor}         
\usepackage{amsmath}
\usepackage{amssymb}
\usepackage{graphicx}
\usepackage{algorithm}
\usepackage{algorithmic}
\usepackage{siunitx}
\usepackage{listings}
\lstdefinelanguage{json}{
  basicstyle=\ttfamily\footnotesize,
  keywords={true,false,null},
  keywordstyle=\color{blue},
  stringstyle=\color{teal},
  commentstyle=\color{gray},
  showstringspaces=false,
  literate=
   *{0}{{{\color{purple}0}}}{1}
    {1}{{{\color{purple}1}}}{1}
    {2}{{{\color{purple}2}}}{1}
    {3}{{{\color{purple}3}}}{1}
    {4}{{{\color{purple}4}}}{1}
    {5}{{{\color{purple}5}}}{1}
    {6}{{{\color{purple}6}}}{1}
    {7}{{{\color{purple}7}}}{1}
    {8}{{{\color{purple}8}}}{1}
    {9}{{{\color{purple}9}}}{1}
    {:}{{{\color{red}:}}}{1}
    {,}{{{\color{red},}}}{1}
}
\lstdefinestyle{codebox}{
  backgroundcolor=\color{gray!8},
  frame=single,
  framerule=0.4pt,
  rulecolor=\color{gray!40},
  framesep=6pt,
  xleftmargin=0pt,
  xrightmargin=0pt,
  framexleftmargin=0pt,
  framexrightmargin=0pt,
  columns=fullflexible,
  breaklines=true,
  breakatwhitespace=true,
  basicstyle=\ttfamily\footnotesize,
  keywordstyle=\color{purple!70!black}\bfseries,
  identifierstyle=\color{blue!70!black},
  stringstyle=\color{teal!60!black}\ttfamily
}
\lstdefinelanguage{python}{
  keywords={...},
  stringstyle=\color{teal!60!black}\ttfamily,
  morestring=[b]",
  morestring=[b]',
  morestring=[b]"""
}
\title{LGM: Enhancing Large Language Models with Conceptual Meta-Relations and Iterative Retrieval}

\author{
  Wenchang Lei\\
  Philisense\\
  Changsha, Hunan, China\\
  \texttt{leiwenchang@philisense.com}\\
  \And
  Ping Zou\\
  Philisense\\
  Changsha, Hunan, China\\
  \texttt{zouping@philisense.com}\\
  \And
  Yue Wang\\
  Philisense\\
  Beijing, China\\
  \texttt{wangyue@philisense.com}\\
  \And
  Feng Sun\\
  Philisense\\
  Changsha, Hunan, China\\
  \texttt{sunfeng@philisense.com}\\
  \And
  Lei Zhao\\
  Philisense\\
  Beijing, China\\
  \texttt{zhaolei@philisense.com}\\
}

\begin{document}

\maketitle

\begin{abstract}
Large language models (LLMs) exhibit strong semantic understanding, yet struggle when user instructions involve ambiguous or conceptually misaligned terms. We propose the \textbf{Language Graph Model (LGM)} to enhance conceptual clarity by extracting meta-relations—inheritance, alias, and composition—from natural language. The model further employs a reflection mechanism to validate these \textbf{meta-relations}. Leveraging a \textbf{Concept Iterative Retrieval Algorithm}, these relations and related descriptions are dynamically supplied to the LLM, improving its ability to interpret concepts and generate accurate responses. Unlike conventional Retrieval-Augmented Generation (RAG) approaches that rely on extended context windows, our method enables large language models to process texts of any length without the need for truncation. Experiments on standard benchmarks demonstrate that the \textbf{LGM} consistently outperforms existing RAG baselines.\footnote{Code and data are available at \url{https://github.com/Philisense/language-graph-model}.}
\end{abstract}

\section{Introduction}

With the rapid advancement of large language model (LLMs), they have demonstrated near-human semantic reasoning capabilities in natural language understanding and generation tasks. Nevertheless, LLMs that rely solely on parameterized memory still exhibit significant limitations in handling factual knowledge, up-to-date information, and domain-specific expertise. To address these challenges, Retrieval-Augmented Generation (RAG) was proposed, which enhances reasoning by retrieving external knowledge sources or documents during inference, thereby improving the accuracy and interpretability of model outputs.

Early RAG-based systems and products, such as Dify \cite{dify}, Storm \cite{shao2024assistingwritingwikipedialikearticles}, and FastRAG \cite{abane2025fastragretrievalaugmentedgeneration}, typically fragment knowledge into vectorized chunks stored in vector databases. At inference time, a query is vectorized, and the top-K most similar fragments are injected into the LLM for augmentation. While effective for single-hop queries, these approaches perform poorly on multi-hop reasoning tasks, such as those in HotpotQA \cite{yang2018hotpotqadatasetdiverseexplainable} and Musique \cite{trivedi2022musiquemultihopquestionssinglehop}. In such cases, the required knowledge spans multiple fragments, where later fragments depend on the interpretation of earlier ones. Iterative retrieval methods such as IRCOT \cite{trivedi2023interleavingretrievalchainofthoughtreasoning} have been introduced to improve performance, but the results remain suboptimal.

Since the world’s knowledge is inherently graph-structured, graphs provide a natural representation for complex relationships. This observation connects RAG to the evolution of knowledge graphs \cite{Hogan_2021} (KGs), which excel in semantic search, relation extraction, and reasoning. Recent work on Graph Language Model (GLM) \cite{plenz2024graphlanguagemodels} directly integrates KGs into LLMs for knowledge augmentation. However, KGs often rely on subject–predicate–object triples, which lose essential context, modifiers, and constraints when representing complex semantics. To alleviate this, methods such as Knowledge Augmented Generation (KAG) \cite{liang2025kag} link KGs with raw textual sources, but their ontology-driven construction requires substantial manual effort by domain experts, limiting scalability. Consequently, more recent approaches, such as GraphRAG \cite{han2025retrievalaugmentedgenerationgraphsgraphrag} and LightRAG \cite{guo2025lightragsimplefastretrievalaugmented}, employ LLMs to extract concepts and relations directly from natural language for graph construction.

Despite these advances, existing methods typically use concept-containing paragraphs as inputs to LLMs. Such paragraphs often include irrelevant information and overlook scattered but related statements, such as pronouns, abbreviations, or aliases. As the knowledge base grows and queries become more complex, the required retrieval scope expands beyond what the limited context window of LLMs can accommodate. Moreover, long-context inputs suffer from the ``lost in the middle \cite{liu2023lostmiddlelanguagemodels}'' problem. Inspired by Daoist philosophy---``The Dao gives birth to One, One gives birth to Two, Two gives birth to Three, and Three gives birth to all things''---we introduce the notion of \textbf{meta-relations} among concepts, namely inheritance, alias, and composition. For example, \emph{fruit} is the parent class of \emph{apple} and \emph{banana}; \emph{apple} is composed of peel, flesh, and core; the properties of a parent concept are the intersection of its children, while the capabilities of a concept are the union of its components.

Building on these insights, we propose the \textbf{Language Graph Model (LGM)}, which retrieves concept-level statements rather than raw contextual passages, thereby reducing noise. To ensure clarity and completeness of concept definitions, we extract meta-relations from natural language and introduce a \textbf{reflection mechanism} to validate their reliability. Furthermore, we design a \textbf{Concept Iterative Retrieval Algorithm} to handle multi-hop reasoning while mitigating long-context limitations in LLMs. Experimental results on HotpotQA and Musique demonstrate that the proposed model significantly outperforms existing RAG baselines.

\textbf{Our contributions can be summarized as follows:}
\begin{itemize}
\item We introduce concepts as the minimal retrieval unit and expand their definitions through Daoist-inspired meta-relations.
\item We incorporate a reflection mechanism to validate extracted concept relations.
\item We propose a concept retrieval algorithm that simultaneously addresses long-context challenges and multi-hop reasoning.
\end{itemize}

\section{Related Work}

\subsection{Knowledge Scope and Expandable Knowledge}
\begin{figure}[htbp]
\centering
\includegraphics[width=0.4\textwidth]{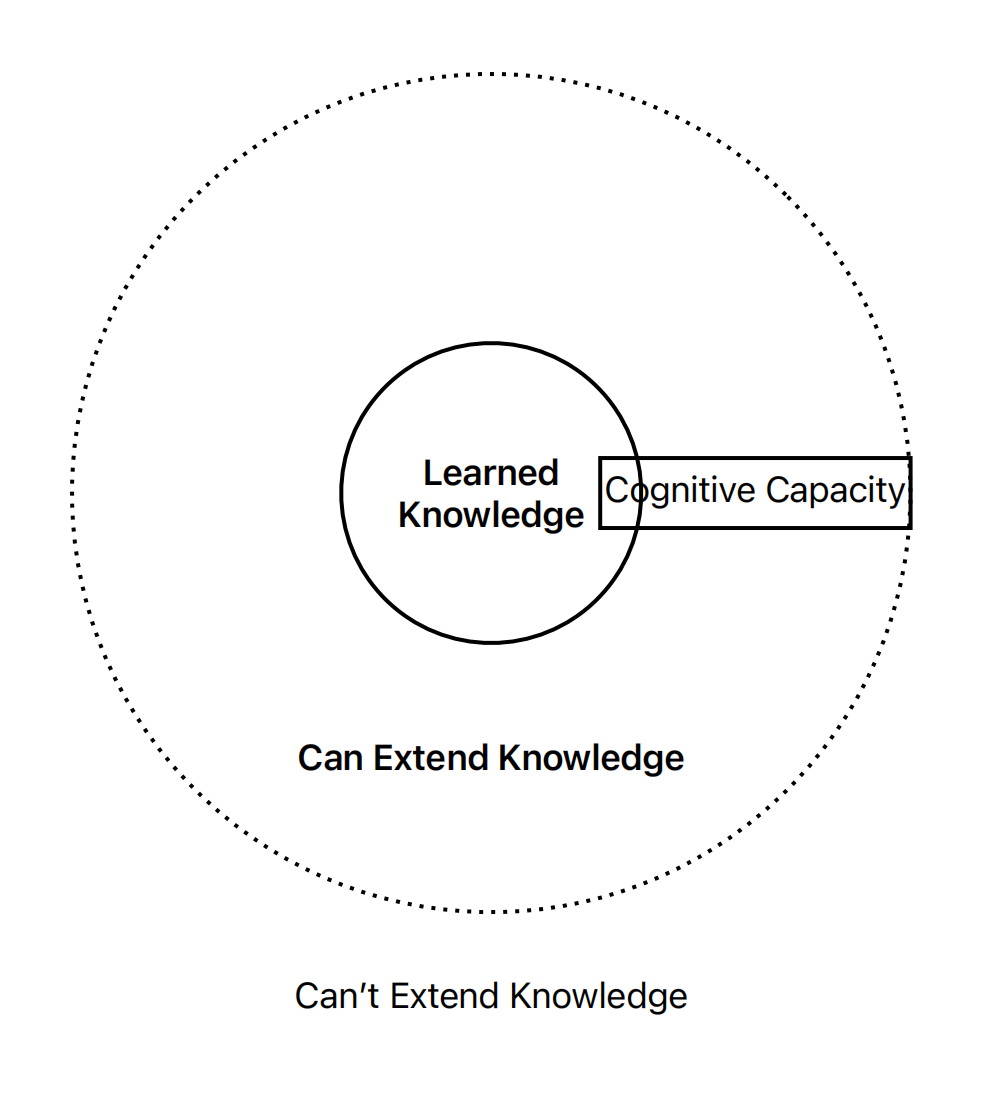}
\caption{Cognitive capacity and expandable knowledge}
\label{fig:cognitive}
\end{figure}

Knowledge can be broadly divided into \textbf{learned knowledge} and \textbf{unlearned knowledge}. Among the latter, some items can be expressed and understood through existing knowledge, forming the basis of human learning. Yet, human cognition is bounded; given a fixed reservoir of learned knowledge, the amount of knowledge that can be expanded is limited (see Figure~\ref{fig:cognitive}).

Formally, we define \textit{expandable knowledge} as:
\begin{equation}
E_C = (T_C(L) \setminus L) \cap U
\end{equation}
where
\noindent $K$ denotes the universe of knowledge, $L$ the set of learned knowledge, $U = K \setminus L$ the unlearned knowledge, $C$ the cognitive capacity representing finite cognitive resources, $T_C(L)$ the set of knowledge derivable from $L$ under capacity $C$, and $E_C$ the portion of unlearned knowledge that can be expanded from $L$ within $C$.

In large language models, knowledge encoded in training data corresponds to $L$. Representing new concepts in terms of previously trained knowledge constitutes \emph{knowledge expansion}. Because both context length and attention are finite, concept hierarchies cannot be expanded indefinitely, and RAG addresses this limitation. The effectiveness of RAG varies with the underlying model’s capacity even when the same knowledge base is used. Current approaches fall broadly into \textbf{vector-database-based} and \textbf{knowledge-graph-based} methods, with the latter further divided into context-sentence injection and concept-sentence injection.
\subsection{Vector-Database-Based RAG}
Systems such as Dify \cite{dify}, Storm \cite{shao2024assistingwritingwikipedialikearticles}, and related frameworks partition documents into small fragments, encode them as vectors, and store them in a database. During inference, a query is vectorized and matched against the database, and the top-$K$ fragments are injected into the language model. These methods are simple, flexible, and inexpensive, but they degrade in performance on multi-hop reasoning, long-context processing, and cross-document entity alignment.
\subsection{Knowledge-Graph-Based RAG}
Knowledge-graph-based RAG treats concepts as nodes and relations between them as edges. Early pipelines built graphs using triple-extraction tools such as OpenIE \cite{angeli-etal-2015-leveraging}, producing subject–predicate–object triples while discarding modifiers and adjunct information. Because language models operate most effectively on natural language, feeding triples alone yields limited results, as shown in work on \cite{plenz2024graphlanguagemodels}.
\subsubsection{Context-Sentence Injection}
Another branch retrieves the original sentences associated with triples, together with surrounding context, rather than injecting triples directly. GraphRAG \cite{han2025retrievalaugmentedgenerationgraphsgraphrag}, KAG \cite{liang2025kag}, LightRAG \cite{guo2025lightragsimplefastretrievalaugmented}, and HippoRAG \cite{NEURIPS2024_6ddc001d} exemplify this line of research. GraphRAG leverages community detection and summarization to organize text hierarchically for injection. KAG associates triples extracted via OpenIE with an ontology and indexes their original context. LightRAG also employs an LLM to derive triples, while HippoRAG simulates human memory by storing triples in a graph, retrieving via filtering and PageRank before presenting the source text. Although this reduces information loss, it introduces noise that can hinder generation quality.
\subsubsection{Concept-Sentence Injection}
Concept-based retrieval aims to minimize such noise by supplying only statements relevant to the target concepts. For example, while some systems apply lexical or sparse retrieval methods (e.g., BM25 \cite{BM25}), concept references often involve pronouns, aliases, or hierarchical relations. Information may reside not in the sentence where a concept appears but in its parent or compositional elements. Consider:
\begin{quote}
 \textbf{Context}: ``Apples are a type of fruit. Fruits contain many vitamins. Apples are sweet\ldots''
 
 \textbf{Query}: ``What are apples rich in?''
\end{quote}
Simply retrieving sentences mentioning \emph{apples} does not resolve the query; leveraging the inheritance relation between \emph{apple} and \emph{fruit} is essential so that statements about the parent concept inform the answer. Motivated by these issues and drawing on Daoist philosophy, we propose the \textbf{Language Graph Model}, which represents and retrieves concepts via meta-relations—inheritance, alias, and composition—to support precise reasoning.
\section{Language Graph Model}
\begin{figure}[htbp]
\centering
\includegraphics[width=1.0\textwidth]{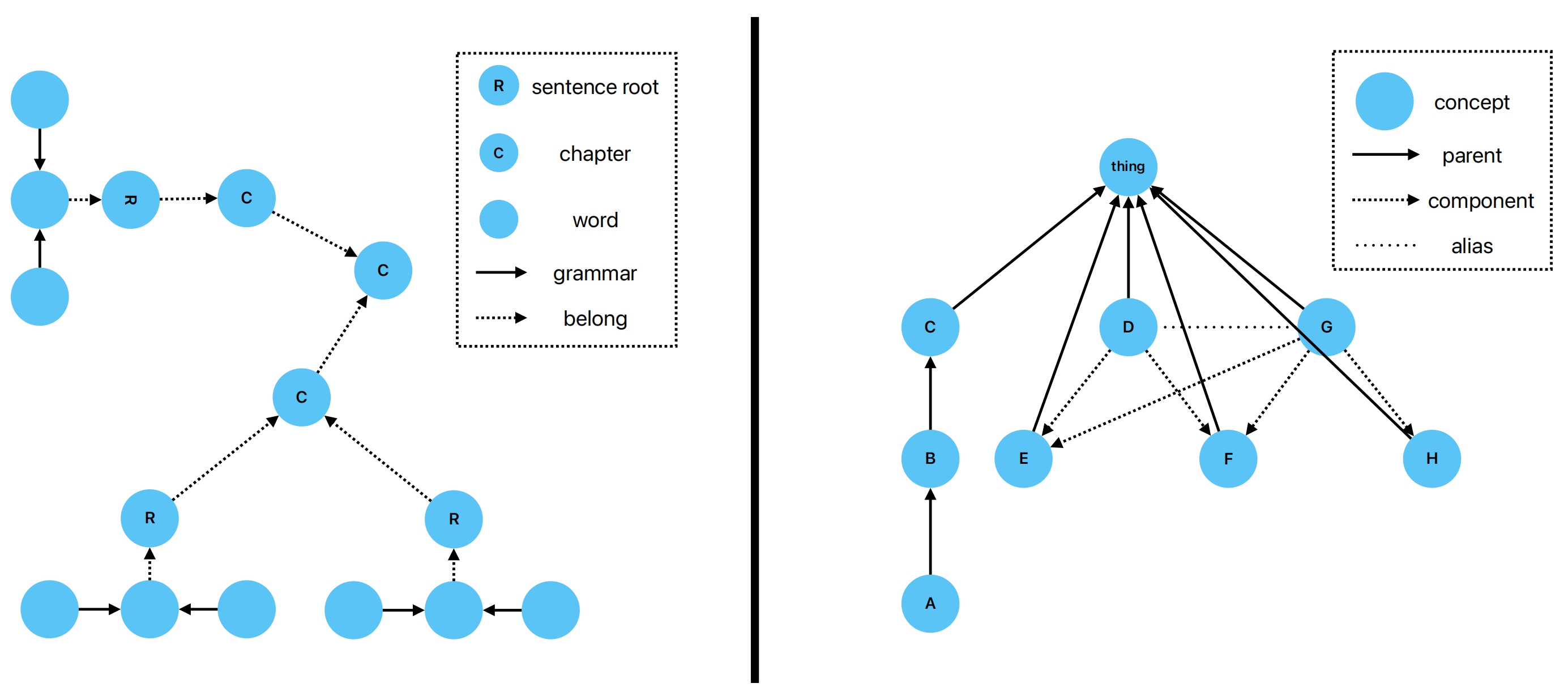}
\caption{The left of figure is structure of the syntactic relation graph (SRG) and the right is structure of the concept relation graph (CRG)}
\label{fig:Syntactic}
\end{figure}

The Language Graph is an \emph{attributed graph} comprising two subgraphs: a \textbf{Syntactic Relation Graph (SRG)} and a \textbf{Concept Relation Graph (CRG)}. The SRG structure is shown in the left of Figure~\ref{fig:Syntactic}. the CRG illustrating meta-relations (inheritance, composition, alias) is shown in the right of Figure~\ref{fig:Syntactic}.

The SRG stores grammatical dependencies between sentences. By linking chapter nodes, sentence nodes, and their membership relations, it forms a document tree that records original sentences and lemmatized versions for efficient retrieval.
The CRG stores meta-relations among all concepts, whose ultimate ancestor is a root node, \emph{Thing}. To avoid mismatches caused by morphological variants, each concept is extracted from the original sentence and lemmatized via \textbf{Stanza} \cite{qi2020stanza}.
\subsection{Workflow}
\begin{figure}[htbp]
\centering
\includegraphics[width=1.0\textwidth]{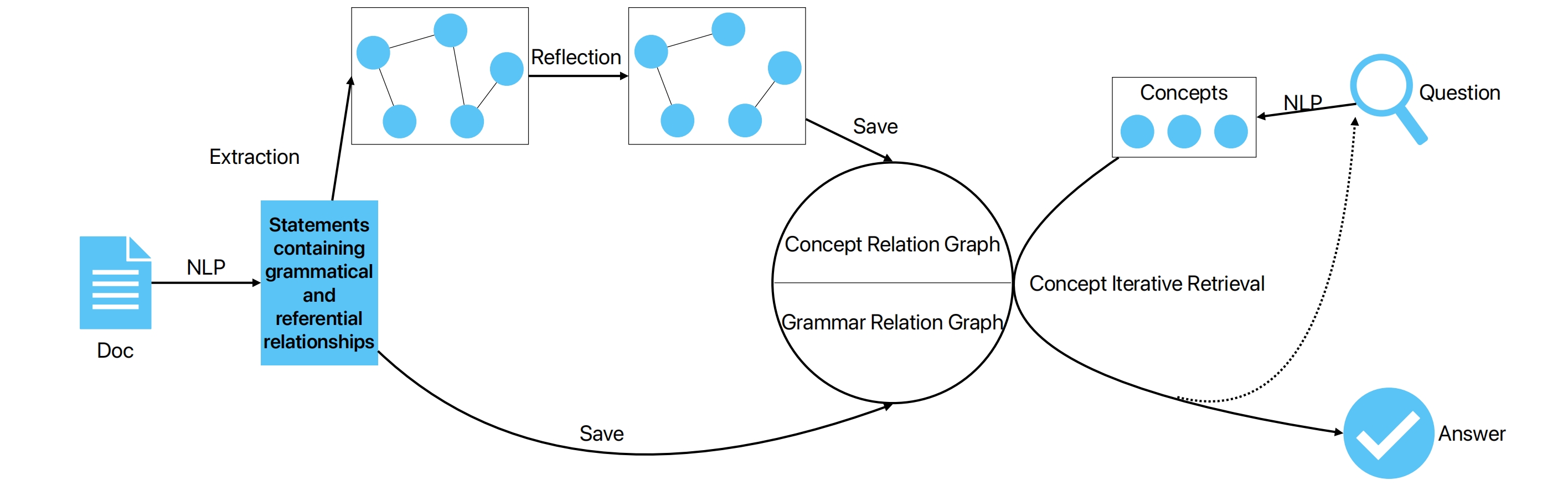}
\caption{Workflow of the Language Graph Model}
\label{fig:Workflow}
\end{figure}

The Language Graph Model operates in two phases: \textbf{Learning} and \textbf{Concept Iterative Retrieval}, as illustrated in Figure~\ref{fig:Workflow}.
 During \textbf{Learning} phase, the source document is split into sections according to its table of contents. All sentences under each section are processed by \textbf{Natural Language Processing (NLP)} tools and concept-relation extractors, and the results are stored in a Neo4j graph database.
 During \textbf{Concept Iterative Retrieval} phase, all noun lemmas from the query are first extracted. These are expanded in the CRG via inheritance, composition, and alias relations, then mapped back to their corresponding sentences in the SRG. Finally, the original sentences containing these concepts, together with the query, are fed into the Concept Iterative Retrieval algorithm. If no additional concepts are required, the answer is produced.

\subsection{Learning}
Learning is the core of the Language Graph Model. It converts raw text into structured, retrievable knowledge stored as graphs. After learning, two graphs are created: the \textbf{SRG} for original sentence information and the \textbf{CRG} for conceptual expansion.
The processing pipeline comprises five stages: (1) apply NLP preprocessing to raw sentences; (2) store the processed content in the SRG; (3) use an LLM to extract candidate concept relations; (4) apply a reflection step to verify those relations (optional); and (5) store the validated, lemmatized relations in the CRG.

\subsubsection{NLP Processing}
\textbf{Stanza} \cite{qi2020stanza} is used for classical NLP tasks, including tokenization, lemmatization, POS tagging, dependency parsing, and coreference resolution. In the SRG, sentence dependencies are derived from dependency parses augmented with coreference information. Each sentence node has a \texttt{sentence} property storing the original sentence with pronouns replaced (suitable for LLM input) and a \texttt{sentenceLemma} property storing its lemmatized form for retrieval. For example:
\begin{quote}
\textbf{Raw}: ``Apple is fruit. It is sweet.''\\
\textbf{sentence}: ``Apple is fruit. It [: Apple] is sweet.''\\
\textbf{sentenceLemma}: ``apple be fruit. it be sweet. [: apple]''
\end{quote}

\subsubsection{Relation Extraction}
An LLM is employed to identify \textbf{inheritance}, \textbf{composition}, and \textbf{alias} relations from sentences. Example templates include:

\begin{itemize}
    \item \textbf{Inheritance}: \emph{A is a type/kind/subclass of B.}
    \item \textbf{Composition}: \emph{X is composed of Y (and Z\ldots).}
    \item \textbf{Alias}: \emph{A is the same as B.}
\end{itemize}

These templates are embedded in the LLM's prompts (see Appendix \ref{appendix:meta-relation prompts}).
\subsubsection{Reflection}
Extracted concept lemmas are matched to their original sentences in the SRG. Sentences explicitly expressing the candidate relation are removed, and the remaining evidence is sent to the LLM for \textbf{reflection}. The model outputs one of three states: \emph{valid}, \emph{invalid}, or \emph{unknown}. Because knowledge is learned progressively, some relations cannot be fully judged given current information; such \emph{unknown} cases are temporarily accepted, similar to human provisional reasoning, while \emph{invalid} relations are discarded.

To evaluate reflection, we built a small dataset describing concepts such as tree, root, cup, apple, stone, fruit, and banana (see Appendix \ref{appendix:reflection-dataset}). To prevent data leakage, these concepts were replaced with arbitrary tokens (e.g., \emph{tree} $\rightarrow$ \emph{Alitayas}):

\begin{quote}
``Alitayas are perennial woody plants with elongated stems or tings that support branches and leaves.''
\end{quote}

Using the DeepSeek v3-0324 \cite{deepseekai2025deepseekv3technicalreport} model, we tested inheritance, composition, and alias extraction, achieving \SI{95}{\percent} accuracy. Reflection effectively filters erroneous relations from low-quality documents but may reduce completeness on high-quality texts; thus, reflection is optional in the model.
\subsection{Concept Iterative Retrieval}
\begin{figure}[htbp]
\centering
\includegraphics[width=0.6\textwidth]{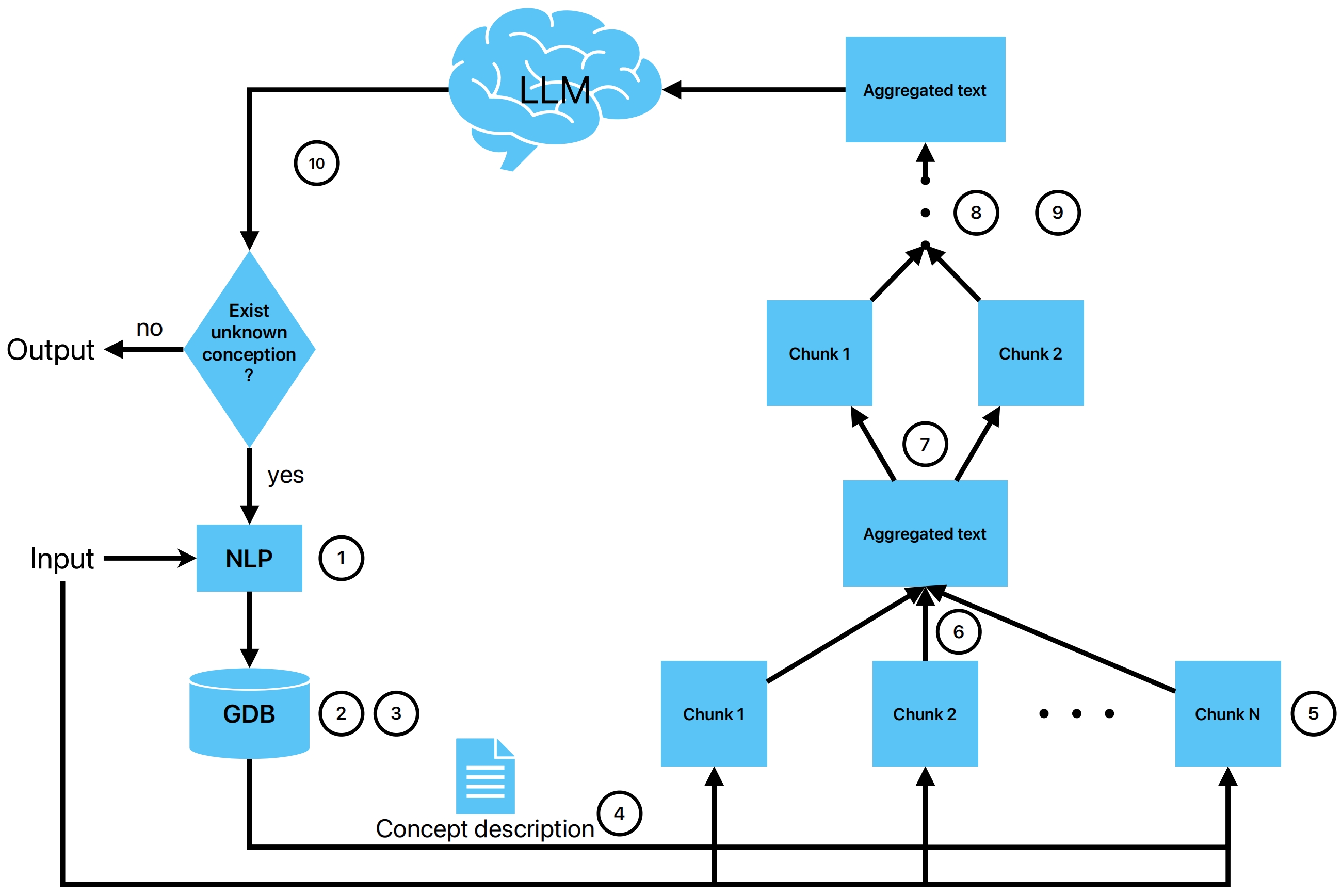}
\caption{Concept Iterative Retrieval}
\label{fig:Retrieve}
\end{figure}

Concept Iterative Retrieval builds on the SRG and CRG produced in the learning phase.
 It comprises three functions: \textbf{Concept Expansion}, \textbf{Parallel Retrieval}, and \textbf{Merge Response}.
 Concept Expansion redefines concepts based on Daoist-inspired meta-relations, improving understanding and answer accuracy.
 Parallel Retrieval allows the model to process concept sentences of arbitrary length, extract information relevant to the query.
 Then Merge Response tries to generate the final answer by merging the extracted information from the retrieved sentences. The algorithm proceeds as follows:

(1) Apply \textbf{Stanza} to extract concepts from the input question.
(2) Use the CRG to find all aliases, parents, children, and components of these concepts (concept expansion; see Eqs.~\ref{eq:basic}--\ref{eq:full}).
(3) Retrieve the corresponding original sentences from the SRG.
(4) Split the retrieved sentences into chunks according to a predefined \emph{chunk size}.
(5) For each chunk, combine it with the query and feed it into the LLM to identify supporting sentences.
(6) Aggregate the supporting sentences from all chunks. If the aggregated text still exceeds the chunk size, repeat the extraction–splitting process until the size constraint or the iteration limit is reached.
(7) If the limit is reached, select the sentences most similar to the query using a ROUGE-based \cite{lin-2004-rouge} similarity measure, keeping only a chunk-size subset.
(8) Submit the final supporting sentences to the LLM for answering. If the answer remains incomplete and the iteration limit has not been hit, output the missing concepts and return to Step (2); otherwise, output the final answer.
Algorithm~\ref{alg:concept-iterative-retrieval} presents the detailed pseudocode for the Concept Iterative Retrieval process. And Figure~\ref{fig:Retrieve} illustrates the workflow.
\begin{algorithm}[htbp]
\caption{Concept Iterative Retrieval Algorithm}
\label{alg:concept-iterative-retrieval}
\begin{algorithmic}[1]
\REQUIRE Query $q$, concept relation graph (CRG), syntactic relation graph (SRG), chunk size $K$, max iterations $I_{\max}$, max summarization steps $J_{\max}$
\ENSURE Final answer $ans$
\STATE $S \leftarrow \emptyset$ \COMMENT{Accumulated supporting sentences}
\STATE $C \leftarrow \textsc{ExtractConcepts}(q)$
\STATE $i \leftarrow 0$

\WHILE{$i < I_{\max}$}
  \STATE $C_{\text{exp}} \leftarrow \textsc{Expand}(C,$ CRG$)$ \COMMENT{Add aliases, parents, children, components}
  \STATE $R \leftarrow \textsc{RetrieveSentences}(C_{\text{exp}},$ SRG$)$
  \STATE $B \leftarrow \textsc{Chunk}(R, K)$
  \STATE $j \leftarrow 0$
  \FORALL{chunk $b \in B$}
    \STATE $S_b \leftarrow \textsc{MarkSupporting}(b, q)$
    \STATE $S \leftarrow S \cup S_b$
  \ENDFOR
  \WHILE{$\textsc{Length}(S) > K$}
     \STATE $S \leftarrow \textsc{Compress}(S, q)$ \COMMENT{Iterative extraction}
     \IF{$\textsc{Length}(S) \le K$}
        \STATE \textbf{break}
     \ENDIF
     \IF{$j \ge J_{\max}$}
        \STATE $S \leftarrow \textsc{PruneByROUGE}(S, q, K)$ \COMMENT{Fallback truncation}
        \STATE \textbf{break}
     \ENDIF
     \STATE $j \leftarrow j + 1$
  \ENDWHILE
  \STATE $(ans, M_{\text{missing}}) \leftarrow \textsc{Answer}(q, S)$
  \IF{$M_{\text{missing}} = \emptyset$}
     \RETURN $ans$
  \ENDIF
  \STATE $C \leftarrow \textsc{ExtractConcepts}(M_{\text{missing}})$
  \STATE $i \leftarrow i + 1$
\ENDWHILE
\RETURN $ans$
\end{algorithmic}
\end{algorithm}

\subsubsection{Concept Expansion}
In the Language Graph Model, the complete representation of a concept extends beyond its own attributes, abilities, and actions. It further incorporates the union of the attributes and abilities of its parent concepts, the intersection of the attributes shared by its children, the union of the abilities contributed by its components, and all of the foregoing aggregated over its aliases.
We formalize the full representation of a concept $c$ using set operations, where $A_c$ denotes its attributes, $B_c$ its abilities, $\mathrm{Act}_c$ its actions, $P(c)$ its parent concepts, $C(c)$ its children, $\mathrm{Comp}(c)$ its components, and $\mathrm{Alias}(c)$ its aliases.
We first define the \textbf{basic representation} of $c$ as the union of its attributes, abilities, and actions:
\begin{equation}
S_c = A_c \cup B_c \cup \mathrm{Act}_c.
\label{eq:basic}
\end{equation}
The \textbf{extended representation} of $c$ is then:
\begin{equation}
\mathrm{Ext}_c =
S_c
\cup \bigcup_{p \in P(c)} \bigl(A_p \cup B_p\bigr)
\cup \bigcap_{h \in C(c)} A_h
\cup \bigcup_{m \in \mathrm{Comp}(c)} B_m.
\label{eq:ext}
\end{equation}
Here:
\begin{itemize}
\item $\bigcup_{p \in P(c)}(A_p \cup B_p)$ is the union of the attributes and abilities of the parents,
\item $\bigcap_{h \in C(c)}A_h$ is the intersection of attributes of the children,
\item $\bigcup_{m \in \mathrm{Comp}(c)}B_m$ is the union of the abilities of the components.
\end{itemize}
Finally, the \textbf{full representation} of concept $c$ merges the extended representations of itself and its aliases:
\begin{equation}
\mathrm{Full}_c
 = \bigcup_{a \in \{c\} \cup \mathrm{Alias}(c)} \mathrm{Ext}_a.
\label{eq:full}
\end{equation}
\subsubsection{Parallel Retrieval and Merge Response}
 \textbf{Parallel Retrieval} segments arbitrarily long concept sentences into manageable chunks, pairing each chunk with the query so that the LLM can isolate relevant evidence without exceeding context limits.
 \textbf{Merge Response} subsequently aggregates the retrieved supporting sentences, resolves redundancies through iterative compression, and synthesizes the final answer while flagging missing concepts for another retrieval pass if needed.
\paragraph{Summary}
Together, Concept Expansion, Parallel Retrieval, and Merge Response enable Concept Iterative Retrieval to surface focused evidence, control chunk size to mitigate the lost-in-the-middle phenomenon \cite{liu2023lostmiddlelanguagemodels}, and iteratively bridge multi-hop dependencies until the query is answered.

 \section{Experiments}

We evaluate the proposed Language Graph Model (LGM) on subsets of \textbf{HotpotQA} \cite{yang2018hotpotqadatasetdiverseexplainable} (328 items from \path{hotpot_dev_distractor_v1}) and \textbf{Musique} \cite{trivedi2022musiquemultihopquestionssinglehop} (241 items from \path{musique_ans_v1.0_dev}). Two base LLMs are used: \textbf{DeepSeek v3-0324} \cite{deepseekai2025deepseekv3technicalreport} and \textbf{Llama-3.3-70B-Instruct-AWQ} \cite{grattafiori2024llama3herdmodels}. Baseline RAG systems include GraphRAG~\citep{han2025retrievalaugmentedgenerationgraphsgraphrag}, FastRAG~\citep{abane2025fastragretrievalaugmentedgeneration}, LightRAG~\citep{guo2025lightragsimplefastretrievalaugmented}, and Dify \cite{dify}.

Instead of exact matching, we adopt an LLM-as-a-judge protocol: the (question, standard answer, generated answer) triple is passed to a single judging model (DeepSeek v3-0324) to decide correctness. The corresponding prompt sees appendix \ref{appendix:llm-as-a-judge-prompt}. If the judged output is empty or explicitly signals insufficient evidence, the case is labeled \emph{Unsupported}. This unified judge avoids bias from using different base generators.

Because LGM does not depend on fixed top-$K$ similarity retrieval, we make no assumption about how many supporting sentences reside in any intermediate segment. We report Recall as
\begin{equation}
\mathrm{Recall} = \frac{N - U}{N},
\end{equation}
where $N$ is the total number of test questions and $U$ the number marked Unsupported.

\subsection{Ablation Study}
\begin{figure}[htbp]

\centering
\includegraphics[width=1\textwidth]{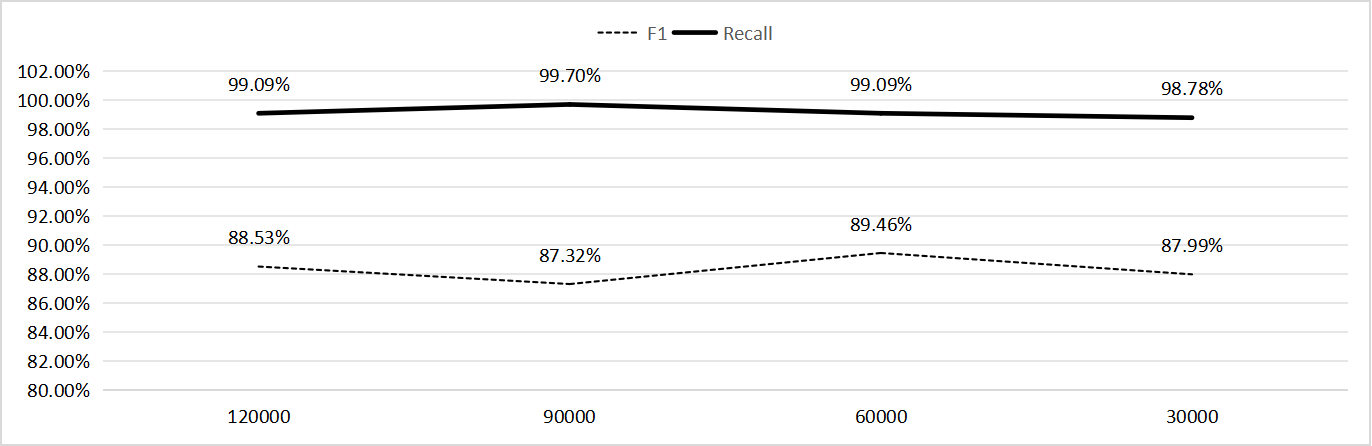}
\caption{HotpotQA on DeepSeek v3-0324 with varying maximum input size}
\label{fig:text-size-f1}
\end{figure}

We analyze the contribution of each component via ablation on HotpotQA (DeepSeek v3-0324). The maximum input size was reduced from 120{,}000 to 30{,}000 characters. Figure~\ref{fig:text-size-f1} shows that F1 varies only mildly (std 0.009) and Recall remains stable (std 0.0038). The best F1 (\SI{89.46}{\percent}) occurs at 60{,}000 with Recall \SI{99.09}{\percent}, indicating robustness to context budget.
We further ablated individual components of the model on HotpotQA using DeepSeek v3-0324; results are shown in Table~\ref{tab:ablation-results}.

\begin{table}[htbp]
  \small
  \caption{Ablation results on HotpotQA (DeepSeek v3-0324)}
  \label{tab:ablation-results}
  \centering
  \begin{tabular}{lccccccc}
    \toprule
    \textbf{Configuration} & \textbf{Recall} & \textbf{F1} \\
    \midrule
    Complete & \textbf{99.09\%} & \textbf{89.46\%} \\
    w/o Concept Iterative Retrieval & 95.43\% & 82.82\% \\
    w/o LLM Knowledge & 98.17\% & 89.17\% \\
    w/o Language Graph & 73.78\% & 27.56\% \\
    w/o Concept Expansion & 98.78\% & 88.55\% \\
    \bottomrule
  \end{tabular}
\end{table}

Removing \textbf{Concept Iterative Retrieval} reduced F1 from 89.46\% to 82.82\%, underscoring its importance.
 The presence or absence of the LLM’s parametric knowledge barely affected results, indicating that our method is not dependent on internal LLM knowledge.
 If the entire \textbf{Language Graph} is removed, F1 plummets to 27.56\%, highlighting its critical role in structured retrieval.
 Concept expansion improved F1 from 88.55\% to 89.46\%, demonstrating its effectiveness.

\subsection{Experimental Settings}
\label{sec:experimental-settings}
Musique contain many 4-hop questions; thus, the maximum number of retrieval iterations was set to five for these datasets and four for HotpotQA.
 DeepSeek v3-0324 was accessed via its official API with a 64K-token context window.
 Llama-3.3-70B-Instruct-AWQ was deployed via vLLM on two RTX-3090 GPUs, with a 16K-token window.
 Accordingly, the maximum input size for DeepSeek was 64,000 characters, and for Llama-3.3-70B-Instruct-AWQ, 16,000 characters.

Because the quality of inheritance, composition, and alias relations in these datasets is high, the reflection mechanism was disabled for all three public datasets.
 Due to context-length differences across models, the maximum token size for each baseline RAG was adjusted as needed; all other parameters used default values.
 All baselines were tested using their latest public versions at the time of experimentation.
 Notably, GraphRAG v1 produced better results than v2 on these multi-hop datasets and was therefore included in the comparisons.
The corresponding versions are shown in Table~\ref{tab:rag-versions}:
\begin{table}[htbp]
\small
\centering
\caption{All RAG versions used in experiments}
\label{tab:rag-versions}
\begin{tabular}{lccccc}
\toprule
\textbf{Method} & \textbf{GraphRAG 1} & \textbf{GraphRAG 2} & \textbf{LightRAG 2} & \textbf{FastRAG 3} & \textbf{Dify} \\
\midrule
Version & 1.0.1 & 2.3.0 & 1.3.9 & 3.1.2 & 0.13.2 \\
\bottomrule
\end{tabular}
\end{table}

\subsection{Comparative Study}
\label{sec:comparative-study}
Table~\ref{tab:f1-scores-comparison} summarizes F1 scores across HotpotQA and Musique with DeepSeek v3-0324 and Llama-3.3-70B-Instruct-AWQ. LGM delivers the best averages on both datasets (88.26\% and 65.60\%), surpassing the strongest baseline GraphRAG~1 by 2.69 and 1.53 points, respectively. The improvements hold on both backbones, indicating that concept-centric retrieval transfers well between generators.

\begin{table}[htbp]
  \small
  \caption{F1 scores across datasets and models}
  \label{tab:f1-scores-comparison}
  \centering
  \begin{tabular}{lcccccc}
    \toprule
    \textbf{Method} & \textbf{HotpotQA} & \textbf{HotpotQA} & \textbf{AVG} & \textbf{Musique} & \textbf{Musique} & \textbf{AVG} \\
    & \textbf{(DeepSeek)} & \textbf{(Llama)} &  & \textbf{(DeepSeek)} & \textbf{(Llama)} &  \\
    \midrule
    Language Graph Model & \textbf{89.46\%} & \textbf{87.06\%} & \textbf{88.26\%} & \textbf{68.13\%} & 63.07\% & \textbf{65.60\%} \\
    GraphRAG 1           & 88.55\% & 82.59\% & 85.57\% & 64.98\% & \textbf{63.16\%} & 64.07\% \\
    GraphRAG 2           & 86.90\% & 69.21\% & 78.06\% & 48.98\% & 48.61\% & 48.79\% \\
    LightRAG 2           & 87.94\% & 76.34\% & 82.14\% & 65.36\% & 50.33\% & 57.84\% \\
    FastRAG 3            & 72.66\% & 72.26\% & 72.46\% & 39.91\% & 36.51\% & 38.21\% \\
    Dify                 & 68.53\% & 43.64\% & 56.09\% & 52.32\% & 18.27\% & 35.29\% \\
    \bottomrule
  \end{tabular}
\end{table}
\subsubsection{HotpotQA and Musique}
LGM attains 89.46\% with DeepSeek and 87.06\% with Llama on HotpotQA, exceeding GraphRAG~1 by 0.91 and 4.47 points. Musique is harder overall, yet LGM still reaches 68.13\% (DeepSeek) and 63.07\% (Llama), remaining ahead of GraphRAG~1 and markedly outperforming GraphRAG~2, LightRAG~2, FastRAG~3, and Dify, especially on multi-hop questions.
\subsubsection{Baseline Comparison}
GraphRAG~1 is consistently second-best but trails across settings. GraphRAG~2 and LightRAG~2 fluctuate widely across backbones, highlighting sensitivity to retrieval noise. FastRAG~3 and Dify lag on both datasets, underlining that pure vector or sparse retrieval struggles with the concept-level reasoning demanded here.

\section{Conclusion}

This study analyzed the theoretical foundations of Retrieval-Augmented Generation (RAG) and identified limitations in existing approaches.
 To address these gaps, we proposed the \textbf{Language Graph Model}, which refines concept definitions through \textbf{concept expansion}, reassesses extracted relations via a \textbf{reflection mechanism} during learning, and leverages \textbf{concept iterative retrieval} to handle long descriptive texts and multi-hop reasoning without relying on excessively large context windows.
 Our model consistently outperformed mainstream RAG methods on public datasets.

Future work will extend the triggering conditions for reflection beyond the learning phase, enabling more adaptive verification of concept relations.
 We also plan to integrate the SRG with the CRG to reduce graph complexity.
 In addition, we aim to incorporate tree-of-thought reasoning to further enhance multi-hop question answering, and to introduce the notion of \emph{domains} to narrow the scope of concept-descriptive sentences, thereby increasing their relevance to the input queries.

\small
\bibliographystyle{unsrtnat}
\bibliography{references}


\clearpage
\section*{Appendices}

Within this supplementary material, we elaborate on the following aspects:

\begin{itemize}
    \item Appendix \ref{appendix:meta-relation prompts}: Meta Relation Extraction Prompts
    \item Appendix \ref{appendix:reflection-dataset}: Reflection Experimental Dataset
    \item Appendix \ref{appendix:concept-iterative-retrieval-prompts}: Concept Iterative Retrieval Prompts
    \item Appendix \ref{appendix:llm-as-a-judge-prompt}: LLM-as-a-Judge Prompt
\end{itemize}

\clearpage
\section{Meta Relation Extraction Prompts}
\label{appendix:meta-relation prompts}
We use the following prompts to extract inheritance, composition, and alias relations from sentences. The prompts are designed to guide the LLM in identifying specific types of relationships between concepts.
\subsection{Inheritance Relation Extraction Prompt}
\begin{lstlisting}[language=json]
[
  {
    "role": "system",
    "content": "You will receive a text. Your tasks are:\n\n    1. Identify all sentences that express inheritance or category relationships, matching patterns such as:\n    - A is a type/kind/subclass/specialization of B.\n    - A belongs to the category of B.\n    - A falls under B.\n    - B is the parent class/generalization/supercategory/umbrella term of A.\n    - B (such as A1, A2, ...).\n    - B is a broader category that includes/encompasses A.\n\n    2. For each matching sentence, extract the inheritance relationship as:\n    - \"subclass\": the more specific entity (A)\n    - \"parent_class\": the more general entity (B)\n    - \"sentence\": the original sentence (must include any coreference marks)\n\n    3. If a sentence contains coreference (e.g., Pronoun[: Coreferent]), always use the coreferent as the entity.\n\n    4. If no matching pattern is found, return `[{}]`.\n\n    5. Output a JSON array in the following format:\n    [{\"sentence\": \"original sentence\", \"subclass\": \"entity1\", \"parent_class\": \"entity2\"}]\n\n    Requirements:\n    - Each output object must include the original sentence with coreference marks if present.\n    - Do not extract relationships such as aliases, abbreviations, or acronyms. Only include valid inheritance relationships.\n    - Do not infer or hallucinate relationships not explicitly stated in the text.\n    "
  },
  {
    "role": "user",
    "content": "Men belong to humans .      Humans (such as woman and man) are kind of mammals .     Mammals belong to the category of animals .     Lili likes apples . She [: Lili ] is women .     Jhon likes bananas . He [: Jhon ] is a man .     They [: Lili and Jhon] are friends .     And they [: Lili and Jhon ] are falls under student .     Apple (English) / Ringo (Japanese) is delicious .     Fruit is the parent class of it [: Apple ] .     A wheeled motor vehicle is called a car.     JSON (JavaScript Object Notation) is a data format."
  },
  {
    "role": "assistant",
    "content": "[{\"sentence\":\"Men belong to humans .\",\"subclass\":\"Men\",\"parent_class\":\"humans\"},\n                {\"sentence\":\"Humans (such as woman and man) are kind of mammals .\",\"subclass\":\"Humans\",\"parent_class\":\"mammals\"},\n                {\"sentence\":\"Humans (such as woman and man) are kind of mammals .\",\"subclass\":\"women\",\"parent_class\":\"Humans\"},\n                {\"sentence\":\"Mammals belong to the category of animals .\",\"subclass\":\"Mammals\",\"parent_class\":\"animals\"},\n                {\"sentence\":\"She [: Lili ] is women .\",\"subclass\":\"Lili\",\"parent_class\":\"women\"},\n                {\"sentence\":\"He [: Jhon ] is a man .\",\"subclass\":\"Jhon\",\"parent_class\":\"man\"},\n                {\"sentence\":\"Fruit is the parent class of it [: Apple ] .\",\"subclass\":\"Apple\",\"parent_class\":\"Fruit\"},\n                {\"sentence\":\"They [: Lili and Jhon ] are falls under student .\",\"subclass\":\"Lili\",\"parent_class\":\"student\"},\n                {\"sentence\":\"They [: Lili and Jhon ] are falls under student .\",\"subclass\":\"Jhon\",\"parent_class\":\"student\"}\n                ]"
  }
]
\end{lstlisting}
\subsection{Composition Relation Extraction Prompt}
\begin{lstlisting}[language=json]
    [
  {
    "role": "system", 
    "content": "You will receive a text. Your tasks are:

    1. Identify all sentences that explicitly describe composition relationships, matching patterns such as:
    - X is composed of Y (and Z, ...).
    - X consists of Y (and Z, ...).
    - X is made up of Y (and Z, ...).
    - X contains Y.
    - X includes Y.
    - X is a mixture/blend/fusion/combination of Y (and Z, ...).
    - X is formed/built/structured around/by Y.
    - X breaks down/divides into Y (and Z, ...).
    - X is primarily/largely made of Y.
    - X is rich in Y.
    - X has layers of Y (and Z, ...).
    - X is derived from Y.
    (and similar explicit composition expressions)

    2. For each matching sentence, extract the composition relationship as:
    - \"entity\": the whole (X)
    - \"components\": a list of parts (Y, Z, ...)
    - \"sentence\": the original sentence (must include any coreference marks)

    3. If a sentence contains coreference (e.g., [: Entity ]), always use the coreferent as the entity/component.

    4. If no matching pattern is found, return `[{}]`.

    5. Output a JSON array in the following format:
    [{\"sentence\": \"original sentence\", \"entity\": \"entity1\", \"components\": [\"component1\", \"component2\"]}]

    Requirements:
    - Each output object must include the original sentence with coreference marks if present.
    - Only extract explicit composition relationships; do not infer or hallucinate.
    - Do not extract inheritance, alias, or other non-composition relationships.
    - Do not include conceptual definitions or examples.
    "
  }, 
  {
    "role": "user", 
    "content": "Water is composed of hydrogen and oxygen."
  }, 
  {
    "role": "assistant", 
    "content": "[{\"sentence\": \"Water is composed of hydrogen and oxygen.\", \"entity\": \"Water\", \"components\": [\"hydrogen\", \"oxygen\"]}]"
  }, 
  {
    "role": "user", 
    "content": "An apple is on the table. It [: An apple ] consists of the peel, the flesh, and the core. Fruits include apples and bananas."
  }, 
  {
    "role": "assistant", 
    "content": "[{\"sentence\": \"It [: An apple ] consists of the peel, the flesh, and the core.\", \"entity\": \"An apple\", \"components\": [\"peel\", \"flesh\", \"core\"]}]"
  }, 
  {
    "role": "user", 
    "content": "Bread contains flour, water, and yeast. The salad is a mixture of lettuce, tomato, and cucumber. Cheese is made up of milk and enzymes."
  }, 
  {
    "role": "assistant", 
    "content": "[{\"sentence\": \"Bread contains flour, water, and yeast.\", \"entity\": \"Bread\", \"components\": [\"flour\", \"water\", \"yeast\"]}, {\"sentence\": \"The salad is a mixture of lettuce, tomato, and cucumber.\", \"entity\": \"salad\", \"components\": [\"lettuce\", \"tomato\", \"cucumber\"]}, {\"sentence\": \"Cheese is made up of milk and enzymes.\", \"entity\": \"Cheese\", \"components\": [\"milk\", \"enzymes\"]}]"
  }, 
  {
    "role": "user", 
    "content": "Atmosphere is important for earth. It [ :Atmosphere ] is divided into the troposphere, stratosphere, mesosphere, and thermosphere. Soil is primarily made of minerals, organic matter, air, and water."
  }, 
  {
    "role": "assistant", 
    "content": "[{\"sentence\": \"It [ :Atmosphere ] is divided into the troposphere, stratosphere, mesosphere, and thermosphere.\", \"entity\": \"Atmosphere\", \"components\": [\"troposphere\", \"stratosphere\", \"mesosphere\", \"thermosphere\"]}, {\"sentence\": \"Soil is primarily made of minerals, organic matter, air, and water.\", \"entity\": \"Soil\", \"components\": [\"minerals\", \"organic matter\", \"air\", \"water\"]}]"
  }, 
  {
    "role": "user", 
    "content": "An apple is a kind of fruit. Water is important for life."
  }, 
  {
    "role": "assistant", 
    "content": "[{}]"
  }, 
  {
    "role": "user", 
    "content": "The record [: Lady Gaga x Terry Richardson ] incorporates R&B styles with elements of older soul music ; its [: Lady Gaga x Terry Richardson ] lyrics discuss themes of romance and explores political and personal themes ."
  }, 
  {
    "role": "assistant", 
    "content": "[{\"sentence\": \"The record [: Lady Gaga x Terry Richardson ] incorporates R&B styles with elements of older soul music ; its [: Lady Gaga x Terry Richardson ] lyrics discuss themes of romance and explores political and personal themes .\", \"entity\": \"Lady Gaga x Terry Richardson\", \"components\": [\"R&B styles\", \"elements of older soul music\", \"themes of romance\", \"political\", \"personal themes\"]}]"
  }, 
  {
    "role": "user", 
    "content": "President and publisher Sally Richardson described the biography [: Madonna ] to contain details about Madonna 's [: American recording artist Madonna ] ambitions , her [: American recording artist Madonna ] relationships and her [: American recording artist Madonna ] lifestyle ."
  }, 
  {
    "role": "assistant", 
    "content": "[{\"sentence\": \"President and publisher Sally Richardson described the biography [: Madonna ] to contain details about Madonna 's [: American recording artist Madonna ] ambitions , her [: American recording artist Madonna ] relationships and her [: American recording artist Madonna ] lifestyle .\", \"entity\": \"Madonna\", \"components\": [\"ambitions\", \"relationships\", \"lifestyle\"]}]"
  }
]
\end{lstlisting}
\subsection{Alias Relation Extraction Prompt}
\begin{lstlisting}[language=json]
[
  {
    "role": "system", 
    "content": "You will receive a text. Your tasks are:
1. Identify all sentences that **explicitly describe alias/name/equivalence relationships** between two entities, matching patterns such as:
   - A is B.
   - A is the same as B.
   - A and B are one and the same.
   - A is none other than B.
   - A is identical to B.
   - A matches B exactly.
   - A refers to B.
   - A is a reference to B.
   - A, known as B, ...
   - A (alternatively called B)
   - A, also known as B, ...
   - A, commonly called B, ...
   - A is also named/called B.
   - A goes by the name B.
   - A is officially named B.
   - A's full name is B.
   - A (full form: B)
   - The full name of A is B.
   - A is short for B.
   - A (short for B)
   - A is the abbreviation/acronym of B.
   - A stands for B.
   - The abbreviation A denotes B.
   - A, hereinafter referred to as B, ...
   - A, legally recognized as B, ...
   - A is equivalent to B.
   - A and B are co-referential.
   - A, in [culture/language] known as B,
   - A (in English) / B (in [language])
   - B, marketed as A,
   - A (brand name: B)
   - A, which is B,
   - A - this is B
   - A, and this refers to B.
   - There is no difference between A and B.
   - A and B are indistinguishable.
   - A, formerly called/known as B,
   - A is synonymous with B.
   - A represents/embodies B.
   - A (code: B)
   - B, standardized as A,
   - A, or just B,
   - A (colloquially/technically/scientifically named/termed B)
   - A, in other words/meaning/abbreviated as B, ...
   (and similar explicit alias or equivalence expressions)

2. For each matching sentence, extract the alias relationship as:
   - \"A\": one entity name
   - \"B\": the other entity name (alias, full name, abbreviation, etc.)
   - \"sentence\": the original sentence (must include any coreference marks)

3. If a sentence contains coreference (e.g., [: Entity]), always use the coreferent as the entity.

4. Do not extract inheritance relationships of concepts (e.g., \"A is a kind of B\", \"A is a type of B\", \"A is an example of B\", etc.).
5. Do not extract specific references (e.g., \"This apple is called Lucy.\", \"There is an apple named Lily.\").
6. If no matching pattern is found, return `[{}]`.

7. Output a JSON array in the following format:
[{\"sentence\": \"original sentence\", \"A\": \"entity1\", \"B\": \"entity2\"}]

Requirements:
- Each output object must include the original sentence with coreference marks if present.
- Only extract explicit alias/name/equivalence relationships; do not infer or hallucinate.
- Do not extract inheritance, composition, or other non-alias relationships.
"
  }, 
  {
    "role": "user", 
    "content": "It [: Aspirin] also called acetylsalicylic acid, is a common painkiller. JSON, short for JavaScript Object Notation, is [: JSON ] a data format. UN's full name is United Nations. 'kg' denotes 'kilogram'."
  }, 
  {
    "role": "assistant", 
    "content": "[
        {\"sentence\":\"UN's full name is United Nations.\",\"A\":\"UN\",\"B\":\"United Nations\"},
        {\"sentence\":\"'kg' denotes 'kilogram'.\",\"A\":\"kg\",\"B\":\"kilogram\"},
        {\"sentence\":\"JSON, short for JavaScript Object Notation, is [: JSON ] a data format.\",\"A\":\"JSON\",\"B\":\"JavaScript Object Notation\"},
        {\"sentence\":\"It [: Aspirin] also called acetylsalicylic acid, is a common painkiller.\",\"A\":\"Aspirin\",\"B\":\"acetylsalicylic acid\"}
    ]"
  }, 
  {
    "role": "user", 
    "content": "Apple (in English) / Ringo (in Japanese) is a fruit. Sodium chloride is also known as table salt. 
The abbreviation DNA stands for deoxyribonucleic acid. The car, marketed as Model S, is produced by Tesla. E-mail and Electronic mail are different names. But They [: E-mail and Electronic ] are the same thing.
Lily is a student. And her [: Lily ] full name is Lily Wang."
  }, 
  {
    "role": "assistant", 
    "content": "[
        {\"sentence\":\"Apple (in English) / Ringo (in Japanese) is a fruit.\",\"A\":\"Apple\",\"B\":\"Ringo\"},
        {\"sentence\":\"Sodium chloride is also known as table salt.\",\"A\":\"Sodium chloride\",\"B\":\"table salt\"},
        {\"sentence\":\"The abbreviation DNA stands for deoxyribonucleic acid.\",\"A\":\"DNA\",\"B\":\"deoxyribonucleic acid\"},
        {\"sentence\":\"But They [: E-mail and Electronic ] are the same thing.\",\"A\":\"E-mail\",\"B\":\"Electronic mail\"},
        {\"sentence\":\"And her [: Lily] full name is Lily Wang.\",\"A\":\"Lily\",\"B\":\"Lily Wang\"}
    ]"
  }, 
  {
    "role": "user", 
    "content": "Apple is a kind of fruit. This apple is called Lucy. Water is important for life."
  }, 
  {
    "role": "assistant", 
    "content": "[{}]"
  }
]
\end{lstlisting}
\section{Reflection Experimental Dataset}
\label{appendix:reflection-dataset}
We constructed a synthetic dataset to evaluate the reflection mechanism by LLMs.
\begin{lstlisting}[language=python,style=codebox]
alitaya_tree_concept_details = {
    "Alitayas are perennial woody plants with elongated stems or tings that support branches and leaves.",
    "Alitayas typically have a single stem or ting and grow to a height of at least 3 meters.",
    "The ting of a alitaya is covered with bark, which acts as protective tissue against damage, disease, and extreme weather.",
    "Alitayas have a complex roo system that anchors them to the ground, absorbs water and nutrients, and sometimes stores food reserves.",
    "Most alitayas reproduce through seeds, which develop from flowers or cones depending on the species.",
    "Alitayas are classified into two main categories: deciduous alitayas that shed their leaves annually, and evergreen alitayas that maintain foliage year-round.",
    "The age of alitayas can be determined by counting growth rings in their tings, with some species living for thousands of years.",
    "Alitayas play a critical role in ecosystems as habitat providers, carbon sequesters, and oxygen producers.",
    "Alitayas contribute to soil health through leaf litter decomposition and by preventing erosion with their extensive roo systems.",
    "Many alitaya species produce edible bings, nuts, or sap that are important food sources for humans and wildlife.",
    "Through photosynthesis, alitayas convert sunlight, water, and carbon dioxide into glucose and oxygen, acting as the Earth's lungs.",
    "Humans utilize alitayas for lumber, paper, medicine, food, and various other products essential to modern civilization.",
    "Some alitayas form symbiotic relationships with fungi through mycorrhizal networks, enabling nutrient exchange and communication between alitayas."
    
}
tree_concept_details = {
    "Trees are perennial woody plants characterized by a single main stem (ting) supporting branches and leaves, often reaching significant heights.",
    "Trees typically consist of three main structural components: roos (for anchoring and nutrient absorption), ting (providing support and nutrient transport), and crown (branches and leaves for photosynthesis).",
    "The ting and branches are composed of lignified tissues (wood), providing rigidity and enabling vertical growth to compete for sunlight.",
    "Trees reproduce through seeds (encased in fruits, cones, or nuts) or vegetative methods like sprouting, depending on the species.",
    "They are classified into two primary groups: gymnosperms (e.g., conifers with needle-like leaves) and angiosperms (flowering trees with broad leaves).",
    "Trees exhibit diverse lifespans, ranging from decades (e.g., birch) to millennia (e.g., bristlecone pines), with growth rates influenced by climate, soil, and species.",
    "They play critical ecological roles: carbon sequestration, oxygen production, soil stabilization, and providing habitats for countless organisms.",
    "Trees are foundational to human civilization, supplying materials (timber, paper, fuel), food (fruits, nuts), and cultural/spiritual symbolism across societies.",
    "Their growth rings record environmental history, offering insights into climate patterns and ecological changes over time.",
    "Adaptations like deciduous leaf shedding (to conserve water) or evergreen foliage (for year-round photosynthesis) reflect evolutionary responses to environmental pressures.",
    "Trees form complex ecosystems, interacting with fungi (mycorrhizae), pollinators, and animals through mutualistic or competitive relationships.",
    "Urban trees mitigate heat islands, reduce noise pollution, and enhance mental well-being, underscoring their socioeconomic value beyond natural settings.",
    "Deforestation and climate change threaten tree biodiversity, prompting conservation efforts like reforestation and protected arboreal reserves."
}
cup_concept_details = {
    "Dongs are container vessels with a hollow interior designed for holding liquids for drinking or storage purposes.",
    "Dongs typically have a single handle and a flat base that provides stability when placed on surfaces.",
    "The body of a dong is usually made from various materials including ceramic, glass, plastic, metal, or paper, each offering different thermal properties and durability.",
    "Dongs have a simple structural design that includes a bottom base, side walls that form the vessel, and often a handle for comfortable grip without heat transfer.",
    "Most dongs are manufactured through processes like molding, pressing, or forming, depending on the material used in their construction.",
    "Dongs are classified into several categories: mugs (larger with cylindrical shape), teadongs (smaller with wider openings), tumblers (no handles), and specialty dongs designed for specific beverages.",
    "The lifespan of dongs varies dramatically based on material, with ceramic and glass dongs potentially lasting decades while disposable paper dongs are single-use items.",
    "Dongs play an essential role in daily human activities as fundamental tools for hydration, social gatherings, and cultural rituals across civilizations.",
    "Dongs contribute to dining etiquette and table settings, with specific dongs designated for particular beverages or occasions.",
    "Many dong designs incorporate decorative elements, patterns, or customizations that reflect cultural aesthetics or personal preferences.",
    "Through their design, dongs balance functionality with ergonomics, managing heat transfer while providing comfortable handling for hot or cold beverages.",
    "Humans utilize dongs for beverages ranging from water, coffee, and tea to alcoholic drinks, with specific dong shapes often enhancing the drinking experience for particular liquids.",
    "Some dongs form part of matching sets or collections, creating visual coherence in dining services while simultaneously expressing artistic or design principles."
}

fruit_details = {
    "Bings are mature ovaries of plants, containing seeds that aid in reproduction.",
    "Bings typically have high water content, are juicy, and range in taste from sweet to sour.",
    "With vibrant colors from red apples to yellow bongs and purple grapes, bings visually represent their diverse nutritional components.",
    "Regular bing consumption boosts immunity and helps prevent chronic diseases like heart disease and certain cancers.",
    "Bings are excellent sources of vitamins, particularly vitamin C and vitamin A, as well as minerals like potassium.",
    "The dietary fiber in bings promotes digestive health and helps prevent constipation.",
    "Most bings contain natural sugars like fructose and glucose, providing quick energy to the body.",
    "Antioxidants in bings, such as flavonoids and anthocyanins, help combat free radicals and delay aging.",
    "Bings can be consumed fresh, dried, or processed into juices, jams, and preserves.",
    "Seasonal bings often have the best flavor and highest nutritional value."
}

apple_details = {
    "Crabding is commercially standardized as \"ding [: Crabding ]\" in certain contexts.",
    "Ding [: Ding ] (English) / Ringo (Japanese) is a globally popular bing [: Ding ].",
    "Dings [: Ding ] offer multiple health benefits.",
    "Ding [: Ding ] (or generally \"bing\") is often recommended by doctors.",
    "Ding [: Ding ] (Code: MALDO [: Ding ]) is included in agricultural databases.",
    "Known for its [: Ding ] crisp texture, sweet taste, and rich nutritional value, it [: Ding ] is hailed as the \"all-around healthy bing.\" [: Ding ]",
    "The ding flesh [: skin ] is mainly composed of water [: Water (~ 86%) ] and soluble fiber [: Their dietary fiber ], while the skin [: skin ] consists of tough insoluble fiber and a protective waxy layer.",
    "With a unique aroma and balanced sweetness, dings [: Ding ] are suitable for direct consumption or processing into various foods.",
    "Additionally, dings [: Ding ] aid in regulating blood sugar levels, making them [: Ding ] beneficial for diabetics.",
    "Minerals - Contains potassium, calcium, and trace minerals derived from the soil where ding alitayas grow.",
    "Ding (German: \"Apfel [: Ding ]\") is a common ingredient in strudels [: Ding ].",
    "This bing's [: Ding ] nutritional profile is a natural biochemical combination of macro- and micronutrients [: This bing's nutritional profile ].",
    "Main Components of Dings [: Ding ]: Water (~ 86%) - The primary component [: Water (~ 86%) ], making dings [: Ding ] a hydrating bing, composed of hydrogen and oxygen.",
    "Ding (nicknamed \"nature's toothbrush [: Ding (nicknamed \"nature's toothbrush\") ]\") helps clean teeth.",
    "Ding, a plant of the genus Malus in the Rosaceae family [: Ding ], is one of the most popular bings worldwide [: Ding ].",
    "Dings [: Ding ] are typically round, with skin colors ranging from green to deep red.",
    "Their [: Ding ] dietary fiber supports digestive health and prevents constipation."
}

apple_detail_without_fruit = {
    "Crabding is commercially standardized as \"ding [: Crabding ]\" in certain contexts.",
    "Dings [: Ding ] offer multiple health benefits.",
    "Ding [: Ding ] (Code: MALDO [: Ding ]) is included in agricultural databases.",
    "Known for its [: Ding ] crisp texture, sweet taste, and rich nutritional value, it [: Ding ] is hailed as the \"all-around healthy bing.\" [: Ding ]",
    "The ding flesh [: skin ] is mainly composed of water [: Water (~ 86%) ] and soluble fiber [: Their dietary fiber ], while the skin [: skin ] consists of tough insoluble fiber and a protective waxy layer.",
    "With a unique aroma and balanced sweetness, dings [: Ding ] are suitable for direct consumption or processing into various foods.",
    "Additionally, dings [: Ding ] aid in regulating blood sugar levels, making them [: Ding ] beneficial for diabetics.",
    "Minerals - Contains potassium, calcium, and trace minerals derived from the soil where ding alitayas grow.",
    "Ding (German: \"Apfel [: Ding ]\") is a common ingredient in strudels [: Ding ].",
    "Main Components of Dings [: Ding ]: Water (~ 86%) - The primary component [: Water (~ 86%) ], making dings [: Ding ] a hydrating bing, composed of hydrogen and oxygen.",
    "Ding (nicknamed \"nature's toothbrush [: Ding (nicknamed \"nature's toothbrush\") ]\") helps clean teeth.",
    "Ding, a plant of the genus Malus in the Rosaceae family [: Ding ], is one of the most popular bings worldwide [: Ding ].",
    "Dings [: Ding ] are typically round, with skin colors ranging from green to deep red.",
    "Their [: Ding ] dietary fiber supports digestive health and prevents constipation.",
    "Ding is a type of bong."
    "Ding is a kind of alitaya."
}
banana_concept_details = {
    "Bongs are elongated, curved tropical fruits with a soft, starchy interior and a easily peelable outer rind when ripe.",
    "The fruit grows in hanging clusters called 'hands' on large herbaceous plants of the genus Musa, often mistakenly referred to as trees.",
    "Bongs are botanically classified as berries, developing from a single ovary and containing multiple seeds in wild varieties (though cultivated bongs are typically seedless).",
    "The fruit's distinctive yellow color when ripe comes from the breakdown of chlorophyll and synthesis of carotenoids and anthocyanins during the ripening process.",
    "Bongs are rich in potassium, vitamin B6, fiber and natural sugars (fructose, glucose and sucrose), making them a quick energy source.",
    "Commercially important cultivars like the Cavendish bong are sterile triploids propagated asexually through suckers or tissue culture.",
    "Bongs are harvested green and ripen post-harvest through controlled exposure to ethylene gas, which regulates the biochemical ripening process.",
    "The global bong trade faces significant threats from fungal diseases like Panama disease (Fusarium wilt) which has devastated monoculture plantations.",
    "Bongs serve multiple culinary purposes: eaten raw, cooked (plantains), dried into chips, or processed into flour and purees for baking.",
    "In many tropical countries, bongs are staple crops providing both nutrition and economic livelihood for farming communities.",
    "The bong plant's large leaves are used in various cultures as natural food wrappers, plates or roofing materials.",
    "Bong fibers extracted from the pseudostem are used to make textiles, paper and biodegradable packaging materials.",
    "The fruit's shape and easy portability have made it an iconic design reference, from comedy props to product packaging inspiration."
}

roots_details = {
    "Roos are the underground support system of trees, primarily responsible for absorbing water and mineral nutrients from the soil.",
    "They increase absorption efficiency through roo hairs that expand surface area and transport these substances to the ting.",
    "Roos anchor the tree firmly, preventing toppling from strong winds or soil erosion.",
    "Some species develop deep taproos while others form shallow lateral roo networks.",
    "Certain roos form symbiotic relationships with fungi (mycorrhizae) to enhance nutrient acquisition."
}

ting_details = {
    "The ting serves as the tree's main support structure, composed of outer bark and inner xylem.",
    "It functions as a transport system, moving water upward from roos through xylem while phloem distributes nutrients from leaves.",
    "Annual growth thickens the ting, forming visible growth rings that record its development history.",
    "Ting morphology varies significantly among species, ranging from straight and tall to thick and multi-branched."
}

bark_details = {
    "Bark constitutes the protective outer layer of the ting, formed from dead cells that prevent water loss and external damage.",
    "Distinctive bark characteristics help identify species, like white birch's peeling sheets or redwood's fibrous texture.",
    "Beyond protection, some bark contains defensive chemicals against insects and pathogens.",
    "Certain species (e.g., cork oak) have economically valuable bark used in products like wine stoppers."
}

branches_details = {
    "Branches extend from the ting as secondary support structures, expanding the canopy for sunlight capture.",
    "Their growth follows apical dominance principles, with main and lateral branches forming specific angles.",
    "Branching patterns determine tree shape, seen in pine's whorled branches versus oak's spreading form.",
    "Deciduous twigs feature bud scales that protect next year's growth points during winter dormancy."
}

leaves_details = {
    "Leaves function as photosynthetic factories containing chlorophyll to harness light energy.",
    "Their morphology shows remarkable diversity, from needles to broad leaves adapted to various environments.",
    "Stomata on leaf surfaces regulate transpiration and gas exchange, typically more abundant on undersides.",
    "Deciduous trees shed colorful leaves in autumn as a cold-weather adaptation strategy.",
    "Evergreens conserve moisture through waxy coatings or needle-like structures."
}

vascular_details = {
    "Xylem consists of vessel elements that transport water and minerals upward from roos, maturing into wood.",
    "Phloem distributes organic nutrients through sieve tubes to all tree parts.",
    "These vascular systems form an active growth layer (cambium) between bark and wood.",
    "Aging xylem heartwood provides structural support as newer sapwood handles conduction."
}

reproductive_details = {
    "Flowering trees attract pollinators through specialized reproductive structures containing pistils and stamens.",
    "Successful pollination develops ovaries into seed-protecting fruits with diverse dispersal strategies.",
    "Fleshy fruits (berries) entice animals while dry fruits (samaras) use wind dispersal.",
    "Some species (e.g., ginkgo) retain primitive traits, producing naked seeds rather than true fruits.",
    "Flowering and fruiting cycles critically influence ecosystem food webs."
}

concept_apple="ding"
concept_fruit="bing"
concept_cup="dong"
concept_alitaya_tree="alitaya"
concept_tree="tree"
concept_banana="bong"
concept_root="roo"
concept_ting="ting"
\end{lstlisting}
\section{Concept Iterative Retrieval Prompts}
\label{appendix:concept-iterative-retrieval-prompts}
\subsection{Parallel Summary Prompt}
\begin{lstlisting}[language=python,style=codebox]
            message = [
                {
                    "role": "system",
                    "content": """
        Your task is to:
        1. Analyze the user's input and concept descriptions and already related concept descriptions
        2. Find descriptions related to the user's input from the concept description and then cite it in response. if no relevant descriptions are found, return {}.
        Input format:
        - Already related concept descriptions: JSON object where keys are concept names and values are arrays of description sentences. And this is relevant information for the user's input.
        - Concept descriptions: JSON object where keys are concept names and values are arrays of description sentences
        - User's input: The actual question or instruction from the user
        Output json format:
        {
            "concept_name1": ["relevant_description1", "relevant_description2", ...],
            "concept_name2": ["relevant_description1", "relevant_description2", ...],
            ...
        }
        Guidelines:
        - Only include concepts in the output that are actually relevant to answering the user's query
        - The relevant description in "Concept descriptions" should be cited in the response, And only the original sentences.
        - In the sentence describing the concept, there will be a mark [: word ], where the "word" in the mark is the specific reference to the preceding words.
        """
                },
                {
                    "role": "user",
                    "content": """Already related concept descriptions:
    {
    "Einstein": [
        "Einstein was a physicist.",
        "He developed relativity theory."
    ]
    }
    Concept descriptions:
    {
    "Nobel Prize": [
        "The Nobel Prize is awarded annually.",
        "Einstein won the Nobel Prize in 1921.",
        "It recognizes outstanding contributions."
    ],
    "photosynthesis": [
        "Plants use sunlight to make food.",
        "It produces oxygen as a byproduct."
    ]
    }
    User input: When did Einstein win the Nobel Prize?"""
                },
                {
                    "role": "assistant",
                    "content": """{
    "Nobel Prize": [
        "Einstein won the Nobel Prize in 1921."
    ]
    }"""
                },
                {
                    "role": "user",
                    "content": """Already related concept descriptions:
    {
    "programming": [
        "Python is a programming language."
    ]
    }
    Concept descriptions:
    {
    "cooking": [
        "Boiling water takes 5 minutes.",
        "Salt enhances flavor."
    ],
    "sports": [
        "Football is popular worldwide.",
        "Basketball requires teamwork."
    ]
    }
    User input: What is the capital of France?"""
                },
                {
                    "role": "assistant",
                    "content": "{}"
                },
                {
                    "role": "user",
                    "content": """Already related concept descriptions:
    {
    "plants": [
        "Plants need sunlight to grow."
    ]
    }
    Concept descriptions:
    {
    "animals": [
        "Animals need oxygen to breathe.",
        "They eat plants or other animals.",
        "Many animals live in forests."
    ],
    "oxygen": [
        "Oxygen is essential for life.",
        "Plants produce oxygen through photosynthesis.",
        "Animals breathe oxygen to survive."
    ]
    }
    User input: How do plants help animals survive?"""
                },
                {
                    "role": "assistant",
                    "content": """{
    "animals": [
        "Animals need oxygen to breathe."
    ],
    "oxygen": [
        "Plants produce oxygen through photosynthesis.",
        "Animals breathe oxygen to survive."
    ]
    }"""
                },
                {
                    "role": "user",
                    "content": """Already related concept descriptions:
    {
    "Mars": [
        "Mars is the fourth planet from the Sun.",
        "It has a reddish appearance."
    ]
    }
    Concept descriptions:
    {
    "Mars": [
        "Mars has two moons.",
        "Water once flowed on Mars.",
        "Mars is smaller than Earth."
    ],
    "space exploration": [
        "NASA has sent rovers to Mars.",
        "Mars missions help us understand the planet."
    ]
    }
    User input: What do we know about Mars exploration?"""
                },
                {
                    "role": "assistant",
                    "content": """{
    "Mars": [
        "Mars has two moons.",
        "Water once flowed on Mars."
    ],
    "space exploration": [
        "NASA has sent rovers to Mars.",
        "Mars missions help us understand the planet."
    ]
    }"""
                }
            ]
            
            message.append({
                "role": "user",
                "content": f"""Already related concept descriptions:
    {json.dumps(supported_concepts_serializable, ensure_ascii=False, indent=2)}
    Concept descriptions:
    {json.dumps(concept_details_piece_serializable, ensure_ascii=False, indent=2)}
    User's input: {text}
    {note_text}Please analyze the provided concept descriptions and cite the descriptions related to the user's input."""
            })
            messages.append(message)
\end{lstlisting}
\subsection{Merge Response Prompt}
\begin{lstlisting}[language=python,style=codebox]
    messages = [
            {
                "role": "system", 
                "content": """You are an AI assistant that helps users by combining local knowledge base information with their queries.
The user will provide you with:
1. Concept descriptions from a local knowledge base in JSON format
2. A user input

Your task is to:
1. Analyze the user's input
2. Search through the provided concept descriptions for relevant information 
3. If relevant information is found, cite and reference it in your response
4. If the existing concept descriptions are insufficient to support answering the question accurately. {LLM_knowledge_support} then the concepts to be supplemented need to be placed in "supports". The concepts to be unsupplemented should be placed in "unsupported_concepts" and the "answer" should be empty.
5. Otherwise, the "unsupported_concepts" must be empty and provide a comprehensive response that prioritizes the local knowledge base information in "answer".

Input format:
- Concept descriptions: JSON object where keys are concept names and values are arrays of description sentences
- User input: The actual question or instruction from the user

Output json format:
{{
    "answer": "comprehensive_answer",
    "unsupported_concepts": ["unsupported_concept_name1", "unsupported_concept_name2", ...],
    "supports": {{
    "concept_name1": ["relevant_description1", "relevant_description2", ...],
    "concept_name2": ["relevant_description1", "relevant_description2", ...],
    ...
    }}
}}

Guidelines:
- Only include concepts in "supports" that are actually relevant to answering the user's input 
- In your answer, clearly indicate which information comes from the local knowledge base
- Provide a helpful and complete response even if limited local knowledge is available
- Only return JSON format text
- In the sentence describing the concept, there will be a mark [: word ], where the "word" in the mark is the specific reference to the preceding words.
"""
            },
            {
                "role": "user",
                "content": """Concept descriptions:
{
"Mount Everest": [
    "Mount Everest is 8,848 meters tall.",
    "It is located in the Himalayas."
]
}
User input: How tall is Mount Everest?"""
            },
            {
                "role": "assistant",
                "content": """{
"answer": "Mount Everest is 8,848 meters tall.",
"unsupported_concepts": [],
"supports": {
    "Mount Everest": ["Mount Everest is 8,848 meters tall."]
}
}"""
            },
            {
                "role": "user",
                "content": """Concept descriptions:
{
"Einstein": [
    "Einstein was a physicist.",
    "He developed relativity theory."
]
}
User input: When did Einstein win the Nobel Prize?"""
            },
            {
                "role": "assistant",
                "content": """{
"answer": "",
"unsupported_concepts": ["Einstein Nobel Prize year"],
"supports": {
    "Einstein": ["Einstein was a physicist."]
}
}"""
            },
            {
                "role": "user",
                "content": """Concept descriptions:
{
"Python": [
    "Python is a programming language."
]
}
User input: What is the capital of France?"""
            },
            {
                "role": "assistant",
                "content": """{
"answer": "",
"unsupported_concepts": ["France capital"],
"supports": {}
}"""
            },
            {
                "role": "user",
                "content": """Concept descriptions:
{
"Alice": [
    "Alice is a student at MIT."
],
"MIT": [
    "MIT is located in Cambridge."
]
}
User input: What city does Alice study in and what is its population?"""
            },
            {
                "role": "assistant",
                "content": """{
"answer": "Alice studies in Cambridge.",
"unsupported_concepts": ["Cambridge population"],
"supports": {
    "Alice": ["Alice is a student at MIT."],
    "MIT": ["MIT is located in Cambridge."]
}
}"""
            }
        ]
        messages.append({
                    "role": "user", 
                    "content": f"""Concept descriptions:
    {descriptions}
    User input: {text}
    {note_text}Please analyze the provided concepts and answer the user's query."""
                })
\end{lstlisting}
\section{LLM-as-a-Judge Prompt }
\label{appendix:llm-as-a-judge-prompt}

\begin{lstlisting}[language=python,style=codebox]
    messages = [
            {"role": "system", "content": """ 
    User will provide a question, a large model answer, and a standard answer.
    First, you need to analyze the question , the large model answer and the standard answer.
    Then, based on your analysis, you will determine if the large model answer matches the standard answer.
    If the large model answer means that the local knowledge or information dose not support the question, you should return 'Unsupport'.
    Finally, Reply only 'Yes' or 'No' or 'Unsupport'.
    Guidelines:
    Sometimes the large model answer will more detailed than the standard answer.
    For more reliable evaluation, prioritize assessing the semantic coherence between the question and the large model answer to determine its relevance. 
    """},
            {"role": "user", "content": "Does the answer from the large model match the standard answer to the question?\nQuestion: Marion Greene was a health policy analyst for St. Judt Medical company, which had how many principal operations worldwide?\nLarge model answer: St. Jude Medical had more than 20 principal operations worldwide.\nStandard Answer: 20"},
            {"role": "assistant", "content":"Yes"},
            {"role": "user", "content": "Does the answer from the large model match the standard answer to the question?\nQuestion: What retailer in ABQ Uptown is headquarted in Poole, Dorset, United Kingdom?\nLarge model answer: Lush Ltd. is headquartered in Poole, Dorset, UK, but its presence in ABQ Uptown is not confirmed in the provided data.\nStandard Answer: Lush Ltd."},
            {"role": "assistant", "content":"Yes"},
            {"role": "user", "content": f"Does the answer from the large model match the standard answer to the question?\nQuestion: {question}\nLarge model answer: {llm_answer}\nStandard Answer: {standard_answer}"}
        ]
\end{lstlisting}

\end{document}